\documentclass[runningheads]{llncs}
\usepackage[T1]{fontenc}
\usepackage{graphicx,verbatim}
\usepackage{booktabs}
\usepackage{amssymb}
\usepackage{xcolor}
\usepackage{amsmath}
\usepackage{multirow}
\usepackage{multicol}
\usepackage{textcomp}
\usepackage[table]{xcolor} 
\usepackage{colortbl}
\usepackage{makecell}
\usepackage{booktabs}
\usepackage{tabularx}
\usepackage{subcaption}
\usepackage{threeparttable}
\usepackage{microtype}
\usepackage{pifont}
\usepackage{marvosym}
\usepackage[colorlinks=true, linkcolor=blue, citecolor=blue, urlcolor=blue]{hyperref}
\begin{document}
\title{Large-Small Model Collaboration for Zero-Shot Surgical Phase Recognition}
%

\author{Yiyi Zhang\inst{1}, Ying Zheng\inst{2}, Wenxin Fan\inst{1}, Yu Zhu\inst{1}, Yuchen Yuan\inst{1}\textsuperscript{(\Letter)}, \\
Litao Zhao\inst{1}\textsuperscript{(\Letter)}, Zheng Li\inst{3}, \and Pheng-Ann Heng\inst{1,4}} 
\authorrunning{Y. Zhang et al.}
\institute{Department of Computer Science and Engineering, The Chinese University of Hong Kong, Hong Kong, China
\and
Department of Electrical and Electronic Engineering, The Hong Kong Polytechnic University, Hong Kong, China
\and
Department of Surgery, The Chinese University of Hong Kong, Hong Kong, China
\and
Institute of Medical Intelligence and XR, The Chinese University of Hong Kong
    \email{ycyuan22@cse.cuhk.edu.hk, litaozhao@cuhk.edu.hk}}

\maketitle              
\begin{abstract}
Task-specific lightweight models for surgical phase recognition excel at capturing temporal dynamics but generalize poorly under domain shift. Conversely, surgical foundation models (FMs) offer superior transferability via large-scale pretraining, yet their lack of explicit temporal modeling often yields temporally inconsistent predictions, leading to degraded performance. To exploit the complementary strengths of both paradigms, we propose \textbf{La}rge-\textbf{S}mall \textbf{T}emporal adaptation (\textbf{LaST}), a novel large-small collaborative framework that enables zero-shot adaptation to unseen clinical domains. In LaST, the FM initiates the pipeline by generating frame-level phase priors that serve as initial weak supervision. To effectively utilize these noisy phase priors, we introduce an iterative temporal refinement scheme that integrates dynamic quality control to filter reliable predictions and dual-model cross-learning to mitigate confirmation bias. Simultaneously, the lightweight model leverages its intrinsic temporal modeling ability to progressively correct inconsistent predictions and enhance overall accuracy across iterations. At the end, a cycle replay strategy is employed to close the loop: the refined, more accurate predictions are utilized as upgraded supervision signals for the subsequent iterations, fostering a self-reinforcing evolution of both label quality and model capability. Extensive experiments demonstrate that LaST achieves robust adaptation to unseen domains for zero-shot surgical phase recognition, outperforming the baseline (PeskaVLP) by 24.85\%-43.17\% in accuracy and even surpassing fully supervised linear probing and several state-of-the-art few-shot approaches. Codes will be released at \url{https://github.com/YIYIZH/LaST}.


\end{abstract}
\section{Introduction}
\label{sec:introduction}

Surgical phase recognition plays a pivotal role in computer-assisted intervention, offering real-time context for anomaly detection and clinical decision support~\cite{jin2021temporal,zhang2025mejo}. Recent progress has demonstrated that task-specific lightweight models can effectively capture complex temporal dependencies within surgical videos, achieving impressive performance when trained on specific procedures or datasets~\cite{gao2021trans,yang2024surgformer}. However, such models often struggle to generalize in real-world clinical environments, where workflows may vary significantly due to differences in operating room layouts, surgeon preferences, institutional protocols, and patient anatomies~\cite{jaspers2025scaling}. Addressing this domain shift typically requires extensive manual re-annotation and model retraining, which is costly, time-consuming, and impractical for widespread deployment. Although SPA~\cite{yuan2025recognizing} has been proposed to alleviate this issue, it still depends on few-shot annotations for adaptation, making it unsuitable for unseen clinical scenarios where labeled data are unavailable.

\begin{figure}[t]
    \centering
    \includegraphics[width=1\linewidth]{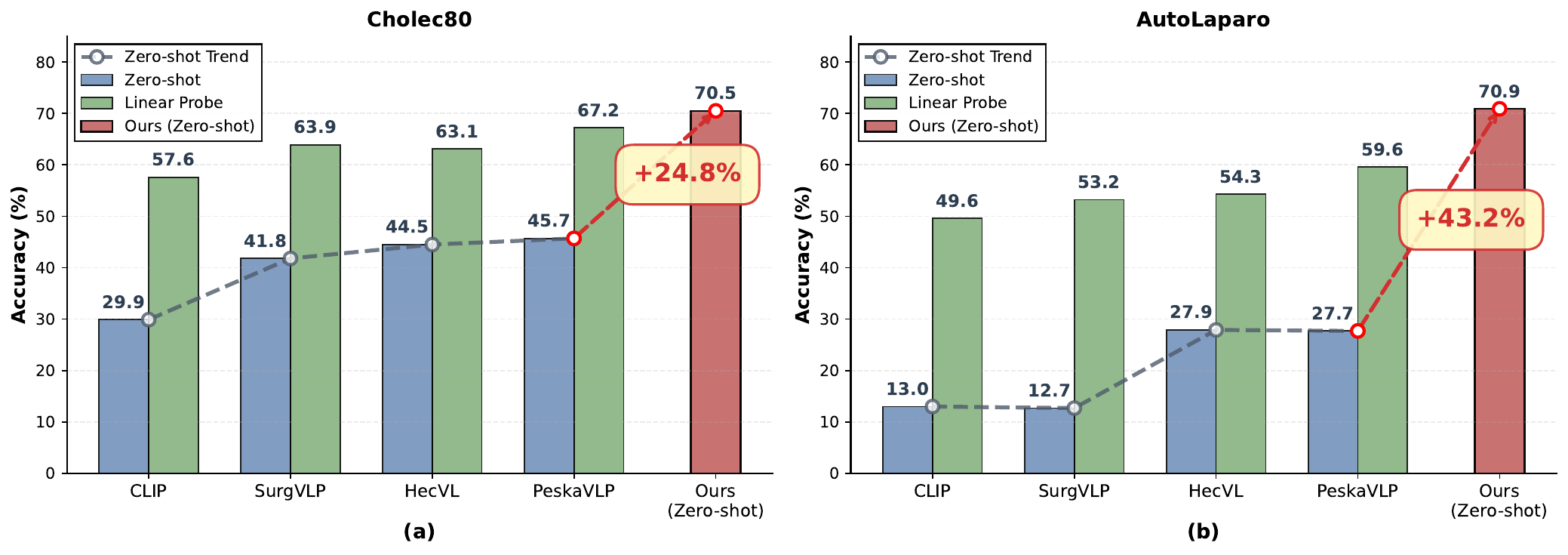}
    \caption{Existing surgical vision-language FMs lack the precision required for clinical utility, and even supervised linear probing remains suboptimal due to the absence of temporal modeling. In contrast, our method addresses these limitations, outperforming PeskaVLP~\cite{yuan2024procedure} by 24.8\% and 43.2\% without manual annotation.}
    \label{fig:teaser1}
\end{figure}
The emergence of surgical vision-language FMs ~\cite{yuan2024procedure,yuan2024hecvl,yuan2025learning} offers a promising direction toward improved generalization.  By aligning visual features with semantic text embeddings, these models demonstrate zero-shot transferability to novel scenarios. However, direct deployment of FMs in clinical settings reveals two fundamental limitations (see Fig. \ref{fig:teaser1}). First, although FMs exhibit strong generalization across domains, their zero-shot predictions often fall short of clinical requirements, lacking the reliability and precision necessary for real-world surgical decision support. Second, fine-tuning approaches, such as linear probe adaptation, are often insufficient to close this gap. This is because most surgical FMs adopt CLIP-like architectures \cite{radford2021learning,zhao2023clip}, which were originally designed for static image-text alignment. As a result, they generally fail to exploit the temporal dynamics inherent in surgical phases \cite{jin2021temporal}, leading to suboptimal performance even after adaptation \cite{shakeri2024few}.

These observations lead to a central question: \textit{Can we design a complementary large-small collaboration in which a foundation model provides broad semantic generalization while a lightweight temporal model captures procedure-specific dynamics, enabling zero-shot adaptation to unseen clinical scenarios?}

An intuitive solution is to leverage an FM to generate initial phase predictions and treat them as weak supervision to train a specialized, temporally aware small model for improved adaptation in surgical phase recognition. However, effective large-small collaboration is non-trivial because FM's zero-shot outputs can be highly noisy and uncertain. Directly distilling these predictions risks causing the small model to overfit unreliable labels, triggering error accumulation and limiting adaptation~\cite{chen2022debiased}. 
Moreover, FM-based label filtering is unreliable across thresholds because FM confidence is not a faithful indicator of prediction quality and can be spuriously high for incorrect predictions. We therefore rely on the small model to absorb noisy supervision, limit error accumulation, and leverage temporal modeling to progressively enhance prediction quality.

To address these challenges, we propose \textbf{La}rge-\textbf{S}mall \textbf{T}emporal adaptation (\textbf{LaST}), a novel large-small collaborative framework that enables zero-shot adaptation to unseen clinical domains for surgical phase recognition. LaST operates through a self-reinforcing strategy comprising three key stages. (1) {Zero-Shot Initialization}: We utilize the FM to generate zero-shot predictions, serving as initial pseudo-labels for the target dataset. (2) {Iterative Temporal Refinement}: We first warm up two identical lightweight temporal models using the FM’s coarse labels. After that, we fit a Gaussian Mixture Model (GMM)~\cite{permuter2006study} to the per-sample loss distribution to distinguish clean from noisy samples in a class-wise manner at each iteration. The two temporal models are then cross-trained on peer-selected ``clean'' subsets, reducing confirmation bias and improving robustness to label noise. (3) {Cycle Replay}: After several iterations of temporal refinement, the refined predictions are fed back as higher-quality pseudo-labels for the next cycle of iterations. Repeating this loop progressively enhances pseudo-label quality and enables the small temporal models to leverage intrinsic temporal modeling capabilities to specialize to the target domain, yielding reliable phase recognition under distribution shift. 

Our main contributions are threefold. (1) We reveal the inherent complementarity between large FMs and lightweight temporal models for surgical phase recognition, and propose a novel large-small collaborative framework (LaST) for zero-shot adaptation. (2) To enable effective large-small collaboration, we design an iterative temporal refinement scheme that integrates a dynamic quality control mechanism and a dual-model cross-learning protocol, together with a cycle replay strategy that forms a self-reinforcing loop. (3) Extensive experiments demonstrate the superior zero-shot adaptation capability of LaST. Specifically, it improves the FM’s F1-score by 23.48\% and 37.54\% on Cholec80~\cite{twinanda2016endonet} and AutoLaparo~\cite{wang2022autolaparo}, respectively, and even surpasses fully supervised linear probing by 1.49\% and 14.61\%. LaST also outperforms the SOTA few-shot adapter, SPA, with 32 shots.

\begin{figure}[t]
    \centering
    \includegraphics[width=1\linewidth]{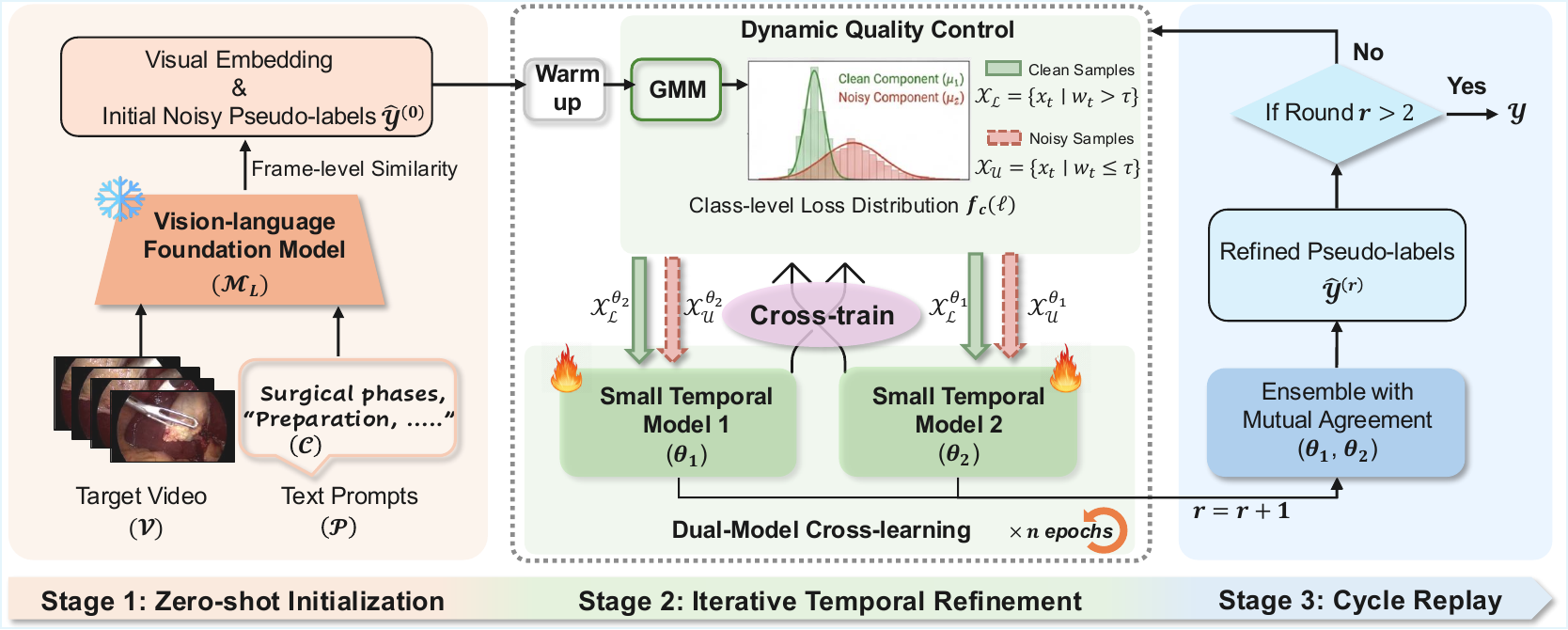}
    \caption{Overview of the LaST framework consisting of three stages. (1) {Zero-Shot Initialization}: the FM generates an initial sequence of pseudo-labels for the target videos; (2) {Iterative Temporal Refinement}: after a warm-up step for dual small models, a dynamic quality control mechanism is used to filter high-quality samples and a cross-learning protocol is adopted to train both models; (3) {Cycle Replay}: the ensemble predictions from the current cycle serve as improved supervision for the next cycle. We take the predictions from the final cycle as the output.}
    \label{fig:main}
\end{figure}

\section{Method}
\label{sec:method}

\subsection{Overview of the LaST Framework}
To effectively adapt a frozen surgical Foundation Model (FM) to unseen target domains without manual annotations, we propose the \textbf{La}rge-\textbf{S}mall \textbf{T}emporal adaptation (\textbf{LaST}) framework, as illustrated in Fig.~\ref{fig:main}. LaST establishes a self-reinforcing loop that synergizes the generalization capabilities of large FMs with the temporal modeling efficiency of lightweight networks. 
Let $\mathcal{V} = \{x_1, x_2, \dots, x_T\}$ denote an unannotated surgical video sequence of length $T$. Our objective is to predict the corresponding sequence of phase labels $Y = \{y_1, y_2, \dots, y_T\}$, where $y_t \in \mathcal{C}$ and $C = |\mathcal{C}|$ represents the number of surgical phase categories.

\subsection{Zero-Shot Initialization}
We employ a pre-trained surgical vision-language FM, $\mathcal{M}_{L}$, to generate initial pseudo-labels for the target video $\mathcal{V}$. For each frame $x_t \in \mathcal{V}$, $\mathcal{M}_{L}$ computes the similarity between the visual embedding of $x_t$ and the text embeddings of a set of predefined prompts $\mathcal{P}$ describing the surgical phases, denoted as:
\begin{equation}
\hat{y}_t^{(0)} = \arg\max_{c \in \mathcal{C}} \text{Sim}(\mathcal{M}_{L}(x_t, \mathcal{P}_c)).
\end{equation}
Let $\hat{\mathcal{Y}}^{(0)} = \{\hat{y}_{1}^{(0)}, \dots, \hat{y}_{T}^{(0)}\}$ denote the pseudo-label sequence for the entire video. Though $\hat{\mathcal{Y}}^{(0)}$ provides useful global semantic priors, it lacks temporal consistency and contains inherent noise, therefore can be regarded as noisy supervision signals. 

\subsection{Iterative Temporal Refinement}
To avoid overfitting to noisy FM pseudo-labels $\hat{\mathcal{Y}}^{(0)}$, we introduce an iterative temporal refinement scheme. We first warm up two identical temporal models with different initializations using $\hat{\mathcal{Y}}^{(0)}$, then divide samples into ``clean'' and ``noisy'' sets by a {dynamic quality control} mechanism. Each model is updated using the peer model’s filtered clean set through dual-model cross-learning.

\subsubsection{Dynamic Quality Control}
Deep networks typically learn clean samples before memorizing noise~\cite{arazo2019unsupervised,chen2019understanding}. We therefore use the per-sample loss as a proxy for label cleanliness and estimate it in a class-wise manner to reduce class-imbalance bias, where samples from rare classes are more likely to be misclassified as noise.
For a specific class $c$, let $\ell_t$ denote the loss of a sample $x_t$ assigned the pseudo-label $c$. We fit a two-component Gaussian Mixture Model (GMM)~\cite{permuter2006study} to the class-specific loss distribution $f_c(\ell)$:
\begin{equation}
f_c(\ell) = \sum_{k=1}^{2} \pi_k \mathcal{N}(\ell \mid \mu_k, \sigma_k^2),
\end{equation}
where $\pi_k$, $\mu_k$, and $\sigma_k^2$ denote the mixture weight, mean, and variance of component $k$, respectively. We assume the component with the smaller mean corresponds to the clean distribution. For each sample $x_t$, we compute the posterior probability~\cite{jiang2016gmm} $w_t = P(\text{clean} \mid \ell_t)$ and partition the data into a labeled set $\mathcal{X}_L = \{x_t \mid w_t > \tau\}$ and an unlabeled set $\mathcal{X}_U = \{x_t \mid w_t \le \tau\}$, based on a confidence threshold $\tau$.

\subsubsection{Dual-Model Cross-Learning}
To mitigate self-training confirmation bias, we jointly train two temporal networks $\theta_1$ and $\theta_2$ via DivideMix-inspired cross-learning~\cite{lidividemix}. Unlike DivideMix, which targets static images with the costly MixMatch strategy, we omit these operations and instead leverage the sequential nature of surgical videos.
Specifically, model $\theta_1$ is trained using the data partition $(\mathcal{X}_L^{\theta_2}, \mathcal{X}_U^{\theta_2})$ determined by the loss distribution of $\theta_2$, and vice versa. For a given model $\theta_i$ ($i \in \{1,2\}$), the total loss $\mathcal{L}_{total}^{\theta_i}$ comprises three terms:

\textit{1) Sparse Supervision.}
We apply cross-entropy loss only on the clean samples 
$\mathcal{X}_L^{\theta_j}$ identified by the peer network $\theta_j$, where $\hat{y}_t$ denotes the pseudo-label of sample $x_t$ and $p_{\theta_i}(\hat{y}_t \mid x_t)$ represents the predicted probability assigned by $\theta_i$:

\begin{equation}
\mathcal{L}_{\text{sparse}}^{\theta_i}
=
-
\frac{1}{|\mathcal{X}_L^{\theta_j}|}
\sum_{x_t \in \mathcal{X}_L^{\theta_j}}
\log p_{\theta_i}(\hat{y}_t \mid x_t),
\quad j \neq i.
\end{equation}

\textit{2) Temporal Smoothing.}
To encourage temporal continuity and reduce abrupt phase transitions, we enforce prediction consistency between adjacent frames in 
$\mathcal{V}$. 
We use a stop-gradient operator ($\text{sg}$)~\cite{chen2021exploring} to stabilize the target, ensuring the prediction of $t$-th frame $x_t$ is smoothed based on its previous frame $x_{t-1}$:
\begin{equation}
\mathcal{L}_{temp}^{\theta_i}
= \frac{1}{T-1} \sum_{t=2}^{T}
\left\|
\log p_{\theta_i}(\cdot \mid x_t)
-
\text{sg}\!\left[
\log p_{\theta_i}(\cdot \mid x_{t-1})
\right]
\right\|_2^2.
\end{equation}

\textit{3). Class Balance Regularization.}
Under severe label noise, models may degenerate to a trivial solution (e.g., predicting a single dominant class) in order to minimize the temporal loss. To prevent such collapse, we impose a prior distribution constraint that encourages balanced predictions across classes.  
Specifically, we introduce a uniform prior distribution $\phi \in \mathbb{R}^{{C}}$ with $\phi_c = \frac{1}{{C}}$ for all $c$. 
The regularization term is defined as the Kullback–Leibler divergence ($D_{KL}$) ~\cite{hershey2007approximating,tanaka2018joint} between the prior $\phi$ and the empirical prediction $\bar{p}_{\theta_i}$:
\begin{equation}
    \mathcal{L}_{reg}^{\theta_i}
= D_{KL}(\phi \parallel \bar{p}_{\theta_i})
= \sum_{c=1}^{{C}}
\phi_c
\log \frac{\phi_c}{\bar{p}_{\theta_i,c}}, \quad \text{where } \bar{p}_{\theta_i}
= \frac{1}{|\mathcal{V}|}
\sum_{x \in \mathcal{V}}
p_{\theta_i}(\cdot \mid x).
\end{equation}

The final objective for each network is:
\begin{equation}
    \mathcal{L}_{total}^{\theta_i} = \mathcal{L}_{sparse}^{\theta_i} +  \mathcal{L}_{temp}^{\theta_i}+  \mathcal{L}_{reg}^{\theta_i}. 
\end{equation}

\subsection{Cycle Replay}
Building upon the intra-cycle refinement, we introduce an outer-loop cycle replay strategy to progressively elevate pseudo-label quality. At the end of each cycle, we obtain refined pseudo-labels $\hat{\mathcal{Y}}^{(r)}$ by ensembling the predictions of $\theta_1$ and $\theta_2$. The refined sequence $\hat{\mathcal{Y}}^{(r)}$ serves as the supervision for the next cycle, replacing the previous $\hat{\mathcal{Y}}^{(r-1)}$. This creates a self-reinforcing loop wherein the models become more accurate across cycles, enabling the quality control mechanism to identify a larger and more reliable set of clean samples, which in turn leads to better models. After three rounds of cycles ($r=3$), we take the ensemble predictions from the final cycle as the final phase output $\mathcal{Y}$ for the target video.

\section{Experimental Setup}
\label{sec:experiments}

\subsubsection{Evaluation Benchmarks:} 
We evaluate our framework on two public benchmarks: \textit{{Cholec80}}~\cite{twinanda2016endonet} and \textit{{AutoLaparo}}~\cite{wang2022autolaparo}. 
{Cholec80} comprises 80 videos of cholecystectomy surgeries, while {AutoLaparo} contains 21 videos of laparoscopic hysterectomy. Both datasets are annotated with 7 distinct surgical phases. We adhere to the official data split protocols for both benchmarks~\cite{twinanda2016endonet,wang2022autolaparo}. We utilize only the testing set videos for both adaptation and final evaluation, ensuring no supervision from any human annotation is accessed.

\textbf{{Implementation Details:}} The proposed LaST framework is implemented using a single RTX 3090. Following SPA~\cite{yuan2025recognizing}, we utilize the frozen visual encoder from PeskaVLP~\cite{yuan2024procedure} as the FM $\mathcal{M}_L$ for our zero-shot initialization; PeskaVLP was pretrained on surgical video lectures (SVL)~\cite{yuan2025learning} from open e-learning platforms. The dual small models ($\theta_1, \theta_2$) are instantiated as MS-TCN~\cite{farha2019ms}. 
All models are optimized using the Adam optimizer with a learning rate of $5 \times 10^{-4}$. The dynamic quality control mechanism performs more reliably with relatively high confidence thresholds ($\tau$>0.7), where $\tau$ = 0.9 achieves the best performance based on sensitivity analysis. We train the framework for 50, 30, and 20 epochs across $r=3$ rounds of cycles, respectively. The training process is lightweight, taking only 0.71 and 0.22 GPU hours on Cholec80 and AutoLaparo.

\begin{table*}[t]
  \centering
  \caption{Surgical phase recognition results on Cholec80 and AutoLaparo datasets. We report Accuracy (Acc) and average F1-Score (F1). The values in subscripts indicate the performance gap compared to LaST. \textbf{Best} and \underline{Second Best}.}
  \setlength{\tabcolsep}{4pt} 
  \resizebox{0.95\textwidth}{!}{
    \begin{tabular}{l c c cc cc}
    \toprule
    & & & \multicolumn{2}{c}{\textbf{Cholec80}} & \multicolumn{2}{c}{\textbf{AutoLaparo}} \\
    \cmidrule(lr){4-5} \cmidrule(lr){6-7}
    \textbf{Model} & \textbf{Shots} & \textbf{Temporal} & \textbf{Acc (\%)} & \textbf{F1 (\%)} & \textbf{Acc (\%)} & \textbf{F1 (\%)} \\
    \midrule
    \multicolumn{7}{l}{\textit{Fully Supervised}} \\
    \rowcolor{green!10}
    TransVNet~\cite{gao2021trans} & Full & \ding{52} & {89.10} & - & {78.30} & - \\
    Linear Probe~\cite{shakeri2024few} & Full & \ding{56} & 67.18$_{\small (+3.34)}$ & \underline{56.83}$_{\small (+1.49)}$ & 59.64$_{\small (+11.22)}$ & 46.41$_{\small (+14.61)}$ \\
    \midrule
    \multicolumn{7}{l}{\textit{Zero-Shot FMs}} \\
    SurgVLP~\cite{yuan2025learning} & 0 & \ding{56} & 41.78$_{\small (+28.74)}$ & 28.14$_{\small (+30.18)}$ & 12.74$_{\small (+58.12)}$ & 9.43$_{\small (+51.59)}$ \\
    HecVL~\cite{yuan2024hecvl} & 0 & \ding{56} & 44.50$_{\small (+26.02)}$ & 25.71$_{\small (+32.61)}$ & 27.90$_{\small (+42.96)}$ & 23.33$_{\small (+37.69)}$ \\
    PeskaVLP~\cite{yuan2024procedure} & 0 & \ding{56} & 45.67$_{\small (+24.85)}$ & 34.84$_{\small (+23.48)}$ & 27.69$_{\small (+43.17)}$ & 23.48$_{\small (+37.54)}$ \\
    \midrule
    \multicolumn{7}{l}{\textit{Few-Shot Adapters}} \\
    Linear Probe~\cite{shakeri2024few}& 32 & \ding{56} & - & 38.37$_{\small (+19.95)}$ & - & 44.72$_{\small (+16.30)}$ \\
    LP+Text~\cite{shakeri2024few} & 32 & \ding{56} & - & 38.36$_{\small (+19.96)}$ & - & 39.40$_{\small (+21.62)}$ \\
    Tip-Adapter-F~\cite{zhang2021tip} & 32 & \ding{56} & - & 42.05$_{\small (+16.27)}$ & - & 39.75$_{\small (+21.27)}$ \\
    SPA~\cite{yuan2025recognizing} & 1 & \ding{52} & - & 44.53$_{\small (+13.79)}$ & - & 51.48$_{\small (+9.54)}$ \\
    SPA~\cite{yuan2025recognizing} & 32 & \ding{52} & - & {55.69}$_{\small (+2.63)}$ & - & \underline{60.36}$_{\small (+0.66)}$ \\
    \midrule
    \textbf{Ours (LaST)} & \textbf{0} & \textbf{\ding{52}} & \textbf{70.52} & \textbf{58.32} & \textbf{70.86} & \textbf{{61.02}} \\
    \bottomrule
    \end{tabular}}
    
    \label{tab:zeroshot_results}
\end{table*}
\section{Experimental Results}
We compare LaST against three categories of state-of-the-art (SOTA) methods, with results summarized in Table~\ref{tab:zeroshot_results}. 
\textbf{VS. Few-Shot Adapters:}
Despite utilizing {zero} manual annotations, LaST remains highly competitive against SOTA few-shot methods. Notably, our zero-shot approach surpasses the 32-shot SPA in F1-score on both datasets. Furthermore, LaST significantly outperforms the 1-shot SPA variant, achieving improvements of {+13.79\%} and {+9.54\%}. This suggests that our self-reinforcing framework effectively mines latent supervision from the unlabeled video structure, offering a robust alternative to few-shot annotation. 
\textbf{VS. Zero-Shot FMs:} 
On the Cholec80 benchmark, LaST surpasses the strongest FM (PeskaVLP) by a substantial margin of {+24.85\%}. The gains are even more pronounced on the challenging AutoLaparo dataset, where LaST improves upon PeskaVLP by {+43.17\%}. These results underscore the critical limitation of applying foundation models in a naive frame-wise manner and validate the efficacy of our temporal refinement module in rectifying noisy zero-shot predictions.
\textbf{VS. Fully Supervised Approaches:}
While a performance gap exists between zero-shot and fully supervised upper bound (e.g., TransVNet), LaST consistently demonstrates superior efficacy against spatially supervised baselines, such as Linear Probe. This highlights the importance of temporal modeling in our method to capture the necessary procedural dynamics for surgical phase recognition.

\begin{figure}[t]
    \centering
    \includegraphics[width=1\linewidth]{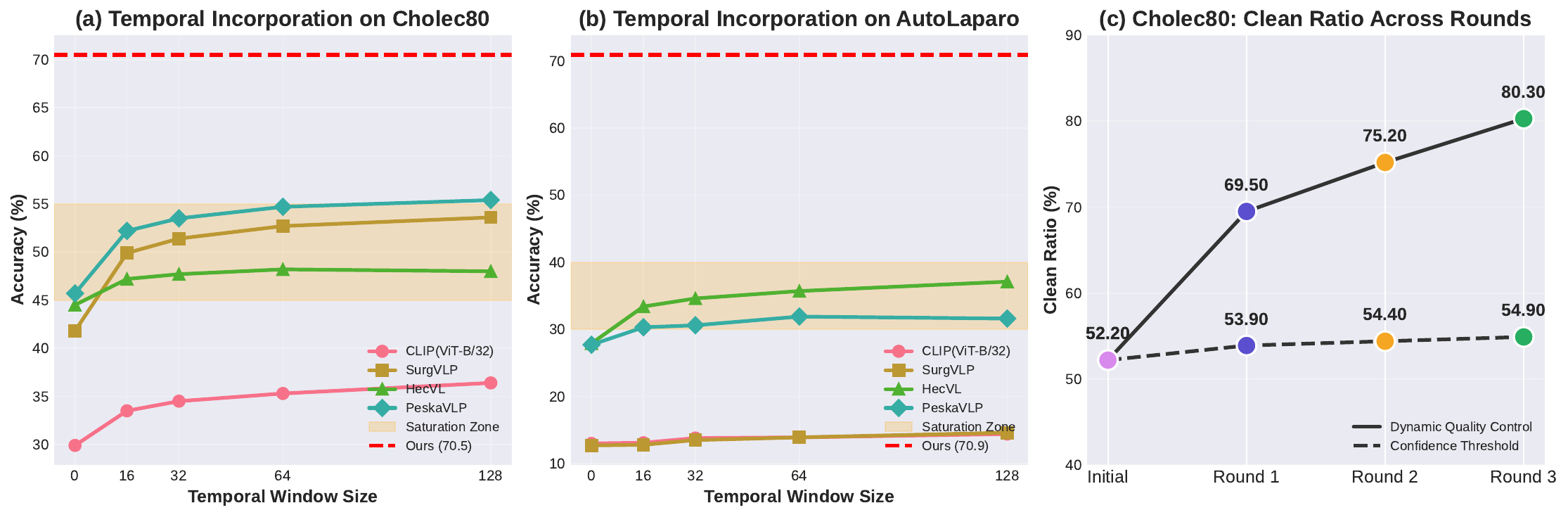}
    \caption{Ablation study on different components of LaST.}
    \label{fig:ablation}
\end{figure}
\begin{table}[t]
\centering
\caption{F1 (\%) across different rounds on AutoLaparo and Cholec80 datasets.}
\setlength{\tabcolsep}{4pt} 
\resizebox{1\textwidth}{!}{
\begin{tabular}{lccc|ccc}
\hline
& \multicolumn{3}{c|}{\textbf{AutoLaparo}} & \multicolumn{3}{c}{\textbf{Cholec80}} \\
\hline
& {Round 1} & {Round 2} & {Round 3} & {Round 1} & {Round 2} & {Round 3} \\
\hline
\textbf{Ours w. single-model} & 50.81 & 56.60 ($+5.79\uparrow$) & {59.72} ($+8.91\uparrow$) & 46.89 &  51.43($+4.54\uparrow$) & {53.80} ($+6.91\uparrow$)\\
\textbf{Ours w. dual-model} & 51.27 & 57.91 ($+6.64\uparrow$) & \textbf{61.02} ($+9.75\uparrow$) & 50.58 & 57.84 ($+7.26\uparrow$) & \textbf{58.32} ($+7.74\uparrow$)\\
\hline
\end{tabular}}

\label{tab:ablation}
\end{table}
\textbf{Ablation Study:}
\textbf{(1)}. We first assess the importance of \textbf{temporal modeling} in the zero-shot pipeline by comparing two strategies for exploiting temporal information: temporal-window feature averaging and MS-TCN. As illustrated in Fig.~\ref{fig:ablation}(a–b), simple averaging over adjacent-frame features consistently improves upon the frame-level baseline (window size 0), indicating that temporal context offers complementary information for surgical phase recognition. Nevertheless, the gains from this naive smoothing scheme quickly plateau, since larger windows may introduce ambiguity from neighboring phases. By contrast, our proposed method effectively overcomes this limitation through adaptive temporal modeling at the video level, yielding a clear advantage over static window-based smoothing.

\textbf{(2)}. We further analyze the effectiveness of our \textbf{dynamic quality control mechanism} based on loss distribution modeling, in comparison with a standard label filtering baseline based on prediction confidence~\cite{liu2025confidence}. To measure the reliability of selected samples, we report the \textit{Clean Ratio}, defined as the percentage of true clean samples among the set of selected ``clean'' pseudo-labels. Results in Fig.~\ref{fig:ablation}(c) highlight a clear divergence between the two strategies. On Cholec80, the baseline quickly saturates, while our method progressively recovers a larger and more accurate clean set over rounds. We observe a similar trend on AutoLaparo, although its curve is omitted for clarity. These results indicate that label filtering based on prediction confidence is unreliable under domain shift, whereas our adaptive loss modeling efficiently separates clean samples from noisy ones.

\textbf{(3)}. We compare the \textbf{dual-model cross-learning} protocol against standard single-model training and assess the effect of \textbf{cycle replay}. The single-model variant removes peer teaching and ensembling, training a single MS-TCN on its own selected labels. As reported in Table~\ref{tab:ablation}, the dual-model variant achieves superior results against single-model (61.02\% VS. 59.72\%, 58.32\% VS. 53.80\%) due to cross-supervision and mutual agreement. In addition, progressively retraining on refined pseudo-labels consistently improves performance, with Round 3 yielding improvements of 9.75\% and 7.74\% over Round 1 with dual-model. These results demonstrate that our proposed iterative temporal refinement scheme and cycle replay strategy are robust to both single and dual modes, where model evolution and pseudo-label quality can mutually reinforce each other across rounds.

\section{Conclusion}
\label{sec:conclusion}

In this work, we presented \textbf{LaST}, a novel framework for zero-shot surgical phase recognition that combines the transferability of surgical FM with the temporal modeling strength of lightweight models to produce accurate predictions under domain shift. By nesting dynamic quality control and dual-model cross-learning within an Iterative Temporal Refinement loop, and further reinforcing this via a global Cycle Replay strategy, our framework enables the system to iteratively self-correct and refine noisy pseudo-labels without human supervision. Experiments show that LaST substantially outperforms zero-shot baselines and even SOTA few-shot methods that require manual annotations, highlighting large-small collaboration as a practical route for deployment in annotation-scarce settings.
    

\begin{credits}

\subsubsection{\ackname} This work was supported by the Research Grants Council of the Hong Kong Special Administrative Region (T45-401/22-N, C4042-23GF, 14214322, 14200623, 14203424, 14206325), the Hong Kong Innovation and Technology Fund (GHP/252/23SZ), the National Natural Science Foundation of China (62106236); and by the AIR@InnoHK Multiscale Medical Robotics Center, the Chow Yuk Ho Technology Center for Innovative Medicine, and the Li Ka Shing Institute of Health Sciences.

\end{credits}

%
%
%
%





\bibliographystyle{splncs04}
\bibliography{bb}

\end{document}